\documentclass[11pt,a4paper]{article}
\usepackage[hyperref]{acl2018}
\usepackage{times}
\usepackage{latexsym}
\usepackage{amsmath}
\usepackage{url}
\usepackage{soul}
\usepackage{graphicx}

\aclfinalcopy 

\title{Natural Language to Code Using Transformers}

\author{\qquad \quad Uday Kusupati \qquad \quad 
  Venkata Ravi Teja Ailavarapu \\
  Dept. of Computer Science \\
  The University of Texas at Austin \\
  {\tt \{uday,avrteja\}@cs.utexas.edu} \\
  }
\date{}

\begin{document}
\maketitle
\begin{abstract}
We tackle the problem of generating code snippets from natural language descriptions using the CoNaLa dataset. We use the self-attention based transformer architecture and show that it performs better than recurrent attention-based encoder decoder. Furthermore, we develop a modified form of back translation and use cycle consistent losses to train the model in an end-to-end fashion. We achieve a BLEU score of 16.99 beating the previously reported baseline of the CoNaLa challenge.
\end{abstract}

\section{Introduction}
Semantic parsing is the problem of mapping natural language sentences to a formal representation like lambda calculus expressions. We work on a specific sub-problem in semantic parsing: producing python code snippets from natural language descriptions. The CoNaLa challenge \cite{yin2018mining} provides a dataset of manually annotated code snippets. For example, `numpy concatenate two arrays `a` and `b` along the second axis' will be mapped to the python code \texttt{print(concatenate((a, b), axis=1))}.

There have been multiple deep learning based approaches to semantic parsing \cite{jia2016data, yin2017syntactic, rabinovich2017abstract, dong2018coarse} using attention-based encoder decoder architectures.
All these approaches use one or more LSTM layers with a suitable attention mechanism as the deep architecture. Transformers \cite{vaswani2017attention} are an alternative to these LSTM based architectures. Transformers have been successfully applied in machine translation beating state-of-the-art LSTM architectures. We explore how transformers perform on the task of semantic parsing using the CoNaLa dataset.

The CoNaLa dataset also provides a large number of mined intent and snippet pairs. Mined intents are natural language descriptions that may not contain any specific reference to variable names or arguments. For example, `numpy concatenate two arrays vertically' would be mapped to \texttt{print(concatenate((a, b), axis=1))}. 
\citet{sennrich2015improving} proposed a back-translation technique to use monolingual corpora effectively in neural machine translation. We explore how we can effectively use these mined examples by employing back-translation on the mined code snippets. Recently, cycle consistency \cite{cycle} constraints have given good results in many tasks in computer vision specifically in tasks that deal with multiple representations of similar semantic value. 


Our main contributions in this paper are as follows:
\begin{enumerate}
    \item We apply the transformer model to the task of converting natural language descriptions to code snippets. We show empirically that the transformer model performs better than a LSTM based encoder decoder model with attention on the CoNaLa dataset. We analyse and compare different configurations of the transformer model.
    \item We adapt the back-translation idea \cite{sennrich2015improving} and propose a way to use the continuous intermediate entities synthesized to train an end-to-end model for this task. We show that this gives a marginal improvement in performance. We also explore cycle consistency losses for the task.
\end{enumerate}

\section{Technical Section}
\subsection{Transformer}
A transformer is similar to many sequence to sequence models in the sense that it contains an encoder and decoder to compress the sentence to an encoding and further generate each token conditioned on previous the previous tokens. But the notion of hidden state that has been instrumental in tasks involved with long sequences is replaced with self attention. The input and output are passed to the encoder and the decoder through an embedding layer. In addition to the word embeddings, the positions of the words are encoded into a positional encoding. This is done so that the positional information is not lost due to our choice of attention over recurrent models.

\begin{figure}
  \centering
  \includegraphics[width=\columnwidth]{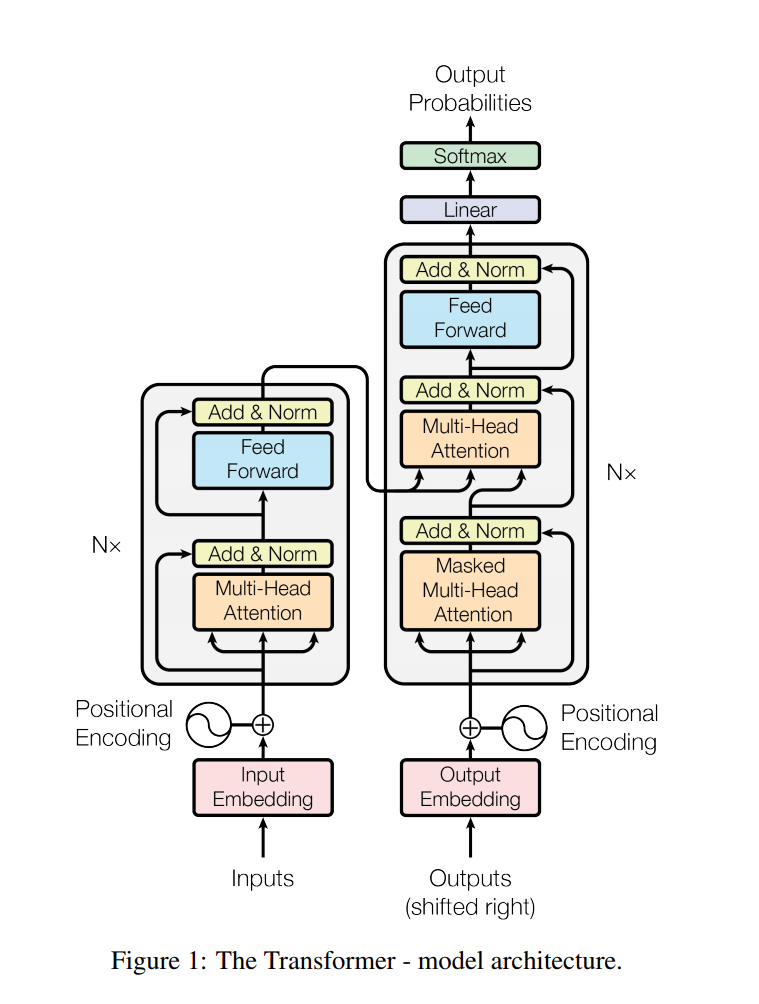}
  \caption{Transformer Architecture. Image Courtesy: \citet{vaswani2017attention}}
  \label{fig:transformer}
\end{figure}

The encoder as well as the decoder consist of stacks of layers which are identical. Each layer generally consists of a multi-head attention along with a feed forward neural network. Attention is computed over the current sequence dealing with, input in the encoder and output in the decoder. It is ensured in the attention layer of the decoder that only the previous words are used to compute self attention specific to each word, which is necessary given the fact that we generate words sequentially during the test time. At the end of a layer, each word vector is passed through a feed forward fully connected network. In addition to this generic setup of self-attention and fully connected network, the decoder layer contains another attention in between them, which computes attention specific to each output word over the input encodings from the output of the encoder stack. 

Since attention is all we use, a single computation of attention may not be sufficient to fetch all the required information from the context. So attention is computed multiple times using different weight matrices, hence giving the name multi-head attention. Batch normalization is used after every layer along with a skip connection from the previous layer. The decoder stacks are connected to a final softmax layer at the end to compute the multinomial distribution over the output vocabulary.


\subsection{Back translation}
We use back translation similar to the procedure described in \citet{sennrich2015improving}. We have a very large mined dataset which has good code snippets but language intents that are not manually curated. So we use this large monolingual code corpus to aid the text to code translation. We do this by using another transformer to back translate from code to text, and using the obtained `text' to reconstruct the code back using our original transformer that translates text to code. In this way, the translation network is forced to learn to predict the code from some encoded representation of the code by the code $\rightarrow$ text transformer. In addition to these back translated `text' the text $\rightarrow$ code transformer is trained on curated data which in turn forces the back translating transformer to perform translation similar to actual text. We perform different experiments on back translation and also on adding noise to the back translated text.
\begin{figure}
  \centering
  \includegraphics[width=\columnwidth]{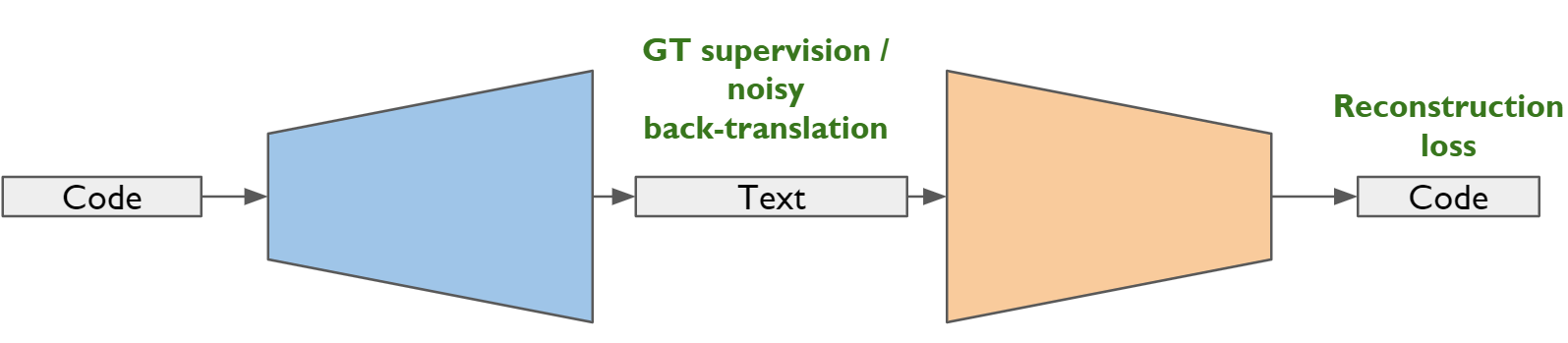}
  \caption{Back translation architecture}
  \label{fig:backtranslation}
\end{figure}
\subsection{Cycle consistency}
We also experiment with an architecture that enforces cycle consistency similar to \citet{cycle}. Since we are using transformers to perform translations in both the directions, code $\rightarrow$ text and text $\rightarrow$ code, we further try to make the translations cycle consistent. So if $F$ represents the translation from text to code and $G$ represents the translation from code to text, we enforce a constraint that,
\begin{align}
    t \in data_{text} \implies G(F(t)) \approx t \nonumber \\
    c \in data_{code} \implies F(G(c)) \approx c \nonumber
\end{align}

\citet{cycle} use Generative Adversarial Networks to model $G$ and $F$ over images from different domains. They do not have domain-paired data instances, hence they use discriminators to train the generators so that they mimic the true distributions, but since we have the true labels for each text-code instance, we just use ground truth supervision on the code and text predictions in conjunction with the cycle consistent loss. 

Unlike images, the intermediate representations $F(t)$ and $G(c)$ are not code and text respectively but multinomial distributions over respective vocabularies. We cannot perform \texttt{argmax} to compute the actual discrete intermediate representation to pass to the next transformer as it would make the computation graph non-differentiable. To overcome this problem we compute the input to the next transformer as follows. Suppose $F(t)$ is represented by a multinomial distribution {$p_i$} over the code vocabulary of size $C$ and the word embedding layer of the code $\rightarrow$ text transformer $G$ is represented by {$e_i$}. $e_i$ is the embedding and $p_i$ is the prediction probability of word $w_i$. Given $p^j_i$ at each position $j$ of the output sequence of transformer $F$, the input code sequence {$c^j$} is computed as,
\begin{align}
    c^j = \sum_i p^j_ie_i \nonumber
\end{align}
This makes the $t \rightarrow F(t) \rightarrow G(F(t)) \approx t$ framework end to end differentiable. This also helps us leverage back translation in an end-to-end manner unlike \citet{sennrich2015improving} where they pretrain the backtranslation network.
\begin{figure}
  \centering
  \includegraphics[width=\columnwidth]{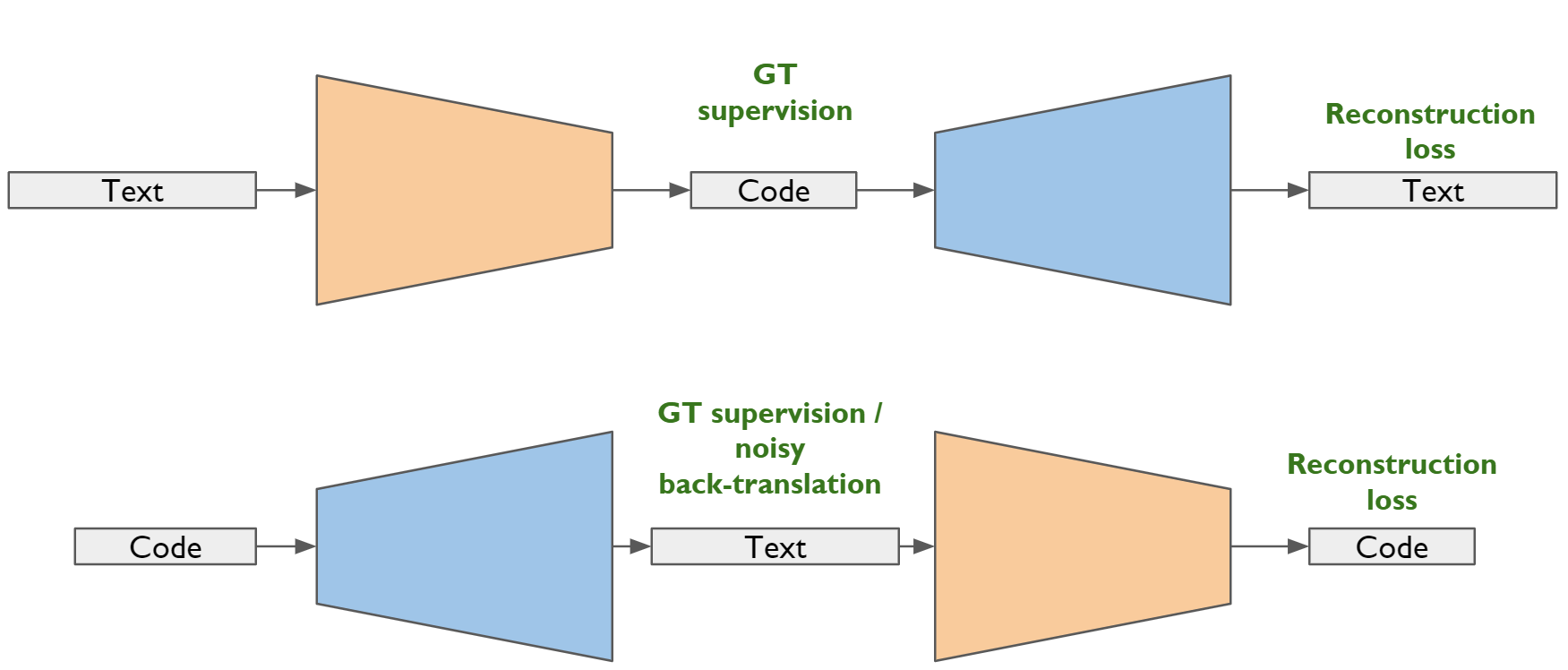}
  \caption{Cycle consistency}
  \label{fig:cycle}
\end{figure}
\section{Experiments}
\subsection{Dataset}

\begin{table*}[t]
    \centering
    \begin{tabular}{c|c|c}
    & Example 1 & Example 2\\
    \hline
    Intent & \emph{trim whitespace} & \emph{Simple way to create matrix of random numbers}\\
    Rewritten Intent & \emph{trim whitespace in string `s`} & \emph{create 3 by 3 matrix of random numbers}\\
    Code Snippet & \texttt{s.strip()} & \texttt{numpy.random.random((3, 3))}
    \end{tabular}
    \caption{Train examples from CoNaLa dataset}
    \label{tab:conala_annot}
\end{table*}

\begin{table*}[h!]
    \centering
    \begin{tabular}{c|c|c}
    & Example 1 & Example 2\\
    \hline
    Intent & \emph{Convert a string to an integer} & \emph{Get a list of values from a list of dictionaries in python}\\
    Code Snippet & \texttt{int('23')} & \texttt{[d['key'] for d in l]}
    \end{tabular}
    \caption{Mined examples from CoNaLa dataset}
    \label{tab:conala_mined}
\end{table*}

We use the dataset released for the CMU CoNaLa, the Code/Natural Language Challenge \cite{yin2018mining}. The dataset consists of two sets of training examples: 2379 annotated examples and about 600,000 mined examples. Tables \ref{tab:conala_annot} and \ref{tab:conala_mined} show some examples from the annotated and mined datasets. The annotated portion of the dataset consists of a small set of examples that are crawled from Stack Overflow, automatically filtered, and curated by annotators. These are split into 2379 training and 500 test examples. Each data instance contains the natural language intent along with the corresponding code snippet. The primary difference between the manually annotated examples and the mined examples is that the natural language intents are rewritten articulating the intent better, typically done by incorporating variable names and function arguments that appeared in the code into the intent.

\subsection{Experimental Details}
We use a transformer with a single layer and 8 attention heads by default.We use cross entropy loss everywhere. We use the Adam optimizer with initial learning rate of 0.001. We use a batch size of 32. We tokenize the intent and snippets using SentencePiece\footnote{\url{https://github.com/google/sentencepiece}}. We set the intent and snippet vocabulary size to be 4000. We use a dropout of 0.2. For inference, we use a beam size of 2. We use the learning rate decay scheme proposed by \citet{vaswani2017attention}. All other hyperparameters are set to default values as reported by \citet{vaswani2017attention}. We conduct all experiments on a single NVIDIA$^{\tiny{\textregistered}}$ GTX 1080 GPU. 

\subsection{Transformer}

\begin{table}[h!]
    \centering
    \begin{tabular}{c|c}
        Model & BLEU Score \\
        \hline
        Encoder Decoder with Attention & 10.58 \\
        Transformer & 16.36
    \end{tabular}
    \caption{Results of the baseline encoder decoder model with attention and our transformer model trained on 2379 annotated examples.}
    \label{tab:baselines}
\end{table}

Table \ref{tab:baselines} shows the results of our single layer transformer model. The results of the baseline attention based encoder decoder model using LSTMs are reported as shown on the CoNaLa challenge website\footnote{\url{https://conala-corpus.github.io/}}. The CoNaLa challenge uses BLEU score as the evaluation metric.


We observe that the transformer model performs significantly better than the baseline model. We look at the predictions of both models on the 500 test examples and analyse why the transformer model performs better. We first observe that the baseline model obtains a BLEU score of zero for about 380 out of the 500 examples while the transformer obtains a BLEU score of zero for about 340 test examples. Though the fraction of zero predictions is high for both models, transformer has lower zero BLEU occurrences. We believe this is one of the primary reasons why the transformer obtains a significantly higher BLEU score.

The dataset contains a diverse set of code snippets in python. Snippets span many python libraries including \texttt{urllib}, \texttt{numpy}, \texttt{os}, \texttt{re}, \texttt{pandas}, and \texttt{datetime}. Though it is diffcult to quantitatively characterize the error classes of each model, we find that the baseline model does well on examples related to the \texttt{datetime} library compared to the transformer. Though the transformer obtains lesser BLEU score on \texttt{datetime} examples, it does much better than the baseline on examples related to the \texttt{pandas}, \texttt{numpy}, and \texttt{re} libraries.  

\begin{table}[h!]
    \centering
    \begin{tabular}{c|c|c}
        Heads & BLEU Score & Token Acc. \\
        \hline
        1 & 14.38 & 52.44 \\
        2 & 15.05 & 54.11 \\
        4 & 15.28 & 55.20 \\
        8 & 16.36 & 56.38 \\
        16 & 14.61 & 56.71
    \end{tabular}
    \caption{Results of the single layer transformer model trained on 2379 annotated examples with different number of attention heads.}
    \label{tab:trans_ablation_heads}
\end{table}

\begin{table}[h!]
    \centering
    \begin{tabular}{c|c|c}
        Layers & BLEU Score & Token Acc. \\
        \hline
        1 & 16.36 & 56.38 \\
        2 & 15.73 & 56.01 \\
        3 & 14.31 & 55.66 \\
        6 & 12.82 & 50.12
    \end{tabular}
    \caption{Results of the transformer model trained on 2379 annotated examples with different number of layers. Number of heads is fixed to be 8.}
    \label{tab:trans_ablation_layers}
\end{table}

To understand the performance of the transformer in greater depth, we perform an ablation on different setting of the number of attention heads and number of layers of the transformer. Table \ref{tab:trans_ablation_heads} shows the results for different number of attention heads. We observe that as the number of heads increase, the performance increases until it plateaus at 8 heads.

Multi head attention is useful in attending over different aspects from the context and thus multiple heads give the network more freedom to learn better features. So the results improve as we increase the heads but start to saturate as we see at 8 heads. This is because of the complexity of the model starting to become much larger compared to the complexity of the problem in hand.

We fixed the number of heads to be 8 and ran the transformer model with different depths. Table \ref{tab:trans_ablation_layers} shows the results of ablation over the number of layers. We observe that the performance of the transformer decreases with the number of layers. This shows that one layer of multi head attention is sufficient to cross the performance of an LSTM based encoder-decoder for this task. This may be due to the fact that the length of the sequences are small.

\subsection{Transformer with Mined Data}
All the results in the previous section were obtained by training the models on the annotated portion of the CoNaLa dataset consisting of 2379 examples. Now we train a five layer transformer using the unannotated mined data too. We try out three different training regimes for using the mined examples along with the annotated train examples. In all cases, we consider the intent annotations in the mined examples as rewritten intent.
\begin{itemize}
    \item \textsc{Mix}: We mix the mined examples with the annotated train examples and train the transformer as before.
    \item \textsc{sample}: We sample equal number of mined and annotated examples during each batch while training. We use a hyperparameter $\alpha$ to weight the loss from the mined examples.
    \item \textsc{finetune}: We first pre-train the transformer using the mined data and then finetune the model using annotated data alone.
\end{itemize}

\begin{table}[h!]
    \centering
    \begin{tabular}{c|c|c}
        Model & BLEU Score & Token Acc. \\
        \hline
        \textsc{Mix} & 13.98 & 60.39 \\
        \textsc{Sample} & 16.57 & 53.62 \\
        \textsc{Finetune} & 16.28 & 58.67 \\
    \end{tabular}
    \caption{Results of the transformer model using different training regimes. In all cases, 100,000 mined examples are used.}
    \label{tab:mined_ablation}
\end{table}

Observing Table \ref{tab:mined_ablation}, we feel that mixing the mined and annotated data performs worse as the mined data doesn't contain rewritten intents and the code snippets in mined data require specific tokens to be copied from the intent. Sampling both data sets equally performs much better as now the annotated data doesn't get lost and their gradients have equal weight. Finetuning appears to be the best as it pre-trains on the huge mined data and then we finetune the weights to the specific task on annotated intent. We also see that there is no correlation between BLEU score and token accuracy. This is because BLEU score also takes into account n-gram precision where as token accuracy is just unigram precision. Furthermore, BLEU score is computed after beam search.

\begin{table}[h!]
    \centering
    \begin{tabular}{c|c}
        Model & BLEU Score \\
        \hline
        Baseline Model (100k) & 14.26 \\
        Transformer (5k) & 15.45 \\
        Transformer (100k) & 16.57 \\
    \end{tabular}
    \caption{Results of the baseline encoder decoder model with attention and our transformer model trained on different amounts of mined data as shown in the brackets.}
    \label{tab:mined_baselines}
\end{table}

Table \ref{tab:mined_baselines} shows the results of the baseline model and our best transformer models when trained using both the annotated and mined data. We observe that adding mined data significantly improves the performance of the baseline encoder decoder model, but we do not observe very significant gains in the performance of the transformer model. We are not able find any good explanation for this trend. 

\subsection{Back Translation and Cycle Consistency}
We train a code $\rightarrow$ text $\rightarrow$ code (\textsc{CTC}) model as follows. In each batch, we sample equal number of annotated and mined examples. The mined examples are passed through both the tranformers while the annotated train examples are only passed through the text $\rightarrow$ code transformer. We also add some gaussian noise to the backtranslated `text' samples before passing them to the text $\rightarrow$ text transformer. We apply a code reconstruction loss on the mined examples and a ground truth supervision loss on the train examples. We use a hyperparameter $\alpha$ to weight the code reconstruction loss. Similarly, we train a text $\rightarrow$ code $\rightarrow$ text (\textsc{TCT}) model.

Further, we train CTC and TCT networks in conjunction to enforce cycle consistency. We use reconstruction loss between the end representations and ground truth supervision loss on the intermediate representation. We sample different batches for each cycle, aggregate the losses and perform backpropagation.

\begin{table}[h!]
    \centering
    \begin{tabular}{c|c|c}
        Model & BLEU Score & Token Acc. \\
        \hline
        \textsc{CTC} & 16.99 & 58.42 \\
        \textsc{CTC-Noise} & 15.65 & 57.89 \\
        \textsc{TCT} & 15.20 & 56.41 \\
        \textsc{Cycle} & 14.04 & 55.89 \\
    \end{tabular}
    \caption{Results of the back-translation based models.}
    \label{tab:back_translation_basic}
\end{table}
Table \ref{tab:back_translation_basic} shows the results of the different back-translation based models that we implemented. We observe that \textsc{CTC} performs the best. Adding noise to the intermediate representations doesn't result in improvements. We also experiment to see how good the reverse argument performs. That is, if we model our task as TCT where text $\rightarrow$ code is the back-translation network rather than the main translation, we still get good results. This shows that the back-translation network is actually learning good intermediate representations that actually are codes, instead of some meaningless intermediate representation just to minimise the reconstruction loss.

We observe that combining both the directions and enforcing cycle consistency does not help. This may be because we also enforce ground truth supervision on text which is irrelevant to our task. Learning a consistent translation in both directions is difficult than learning a good unidirectional translation. It may also be the case that instead of learning to model the distributions better, the model might be just memorizing some translation cycle instances.

\begin{table}[h!]
    \centering
    \begin{tabular}{c|c|c}
        $\alpha$ & BLEU Score & Token Acc. \\
        \hline
        0.0 & 16.57 & 53.62 \\
        0.1 & 16.99 & 58.42 \\
        0.2 & 13.58 & 57.63 \\
        0.5 & 12.84 & 58.21 \\
        1.0 & 12.21 & 57.42 \\
        2.0 & 11.37 & 57.24 \\
        5.0 & 10.63 & 57.16 \\
    \end{tabular}
    \caption{Results of the \textsc{CTC} back-translation based model for different values of the loss weighting parameter $\alpha$.}
    \label{tab:ablate_backtranslation}
\end{table}
Also, we vary the hyperparameter $\alpha$ to find the right balance between reconstruction loss and ground truth supervision. The results are summarized in Table \ref{tab:ablate_backtranslation}. $\alpha=0$ refers to no reconstruction loss which is just the text $\rightarrow$ code transformer (\textsc{Sample}) from Table \ref{tab:mined_ablation}. Better results with small $\alpha$ suggest that though reconstruction loss is useful, ground truth supervision is the main driving force. From the experiments, we choose $\alpha=0.1$.

\section{Related Work}
In this section we discuss some of the works related to our task. 
\subsection{Semantic Parsing}
The task we tackle is a sub-problem of semantic parsing where the target outputs are python code snippets. \citet{jia2016data} proposed an encoder-decoder model with an attention-based copying mechanism for semantic parsing. Some approaches have used syntax of the code language to build better models \cite{yin2017syntactic, rabinovich2017abstract, yin2018tranx}. \citet{dong2018coarse} used a hierarchical approach to neural semantic parsing in which they generated intermediate logical forms to help build the final logical form.

\subsection{Transformers}
In this paper, we primarily use and analyse the transformer model. \citet{vaswani2017attention} introduced the transformer architecture which uses multiple attention heads and a self attention mechanism to compute a hidden vector representation of a sequential input. Transformer architectures have obtained promising results on machine translation tasks \cite{vaswani2017attention} and have recently been successfully applied on language understanding tasks \cite{radford2018improving, devlin2018bert}.

\subsection{Back-translation and Cycle Consistency}
Back translation has been used by \citet{sennrich2015improving} to incorporate large monolingual corpora to aid translation by generating synthetic examples. Similar usage has also been seen in \citet{xie2018noising} to leverage noisy synthetic examples for grammar correction. Cycle consistency has been useful in many tasks in computer vision like unpaired image to image translation, structure from motion, 3D shape matching, cosegmentation, dense semantic alignment \cite{zhou2016learning}, and depth estimation \cite{godard2017unsupervised}.

\section{Conclusion}
In this paper, we showed that the transformer works better than standard attention based recurrent architectures on the task of natural language to code conversion. We experimented with different training regimes to incorporate information from the abundant amount of mined data in the CoNaLa dataset. We devised a modified back-translation strategy to use the mined data in a unsupervised manner. We also explored a cycle consistent loss for the task, but to no avail.

\section*{Acknowledgments}
We thank Prof. Greg Durrett for the helpful discussions and constructive feedback. We also thank Arka Sanka. 

\bibliography{acl2018}
\bibliographystyle{acl_natbib}

\appendix






\end{document}